# Exploring Hardware Friendly Bottleneck Architecture in CNN for Embedded Computing Systems

*Xing Lei, Longjun Liu\*, Zhiheng Zhou, Hongbin Sun, Nanning Zheng*

Institute of Artificial Intelligence and Robotics (IAIR), Xi'an Jiaotong University, Xi'an, China.

\*liulongjun@xjtu.edu.cn

## ABSTRACT

In this paper, we explore how to design lightweight CNN architecture for embedded computing systems. We propose L-Mobilenet model for ZYNQ based hardware platform. L-Mobilenet can adapt well to the hardware computing and accelerating, and its network structure is inspired by the state-of-the-art work of Inception-ResnetV1 and MobilenetV2, which can effectively reduce parameters and delay while maintaining the accuracy of inference. We deploy our L-Mobilenet model to ZYNQ embedded platform for fully evaluating the performance of our design. By measuring in cifar10 and cifar100 datasets, L-Mobilenet model is able to gain 3× speed up and 3.7× fewer parameters than MobileNetV2 while maintaining a similar accuracy. It also can obtain 2× speed up and 1.5× fewer parameters than ShufflenetV2 while maintaining the same accuracy. Experiments show that our network model can obtain better performance because of the special considerations for hardware accelerating and software-hardware co-design strategies in our L-Mobilenet bottleneck architecture.

## 1. INTRODUCTION

Deep learning [1] overcomes some of the problems that were considered difficult to solve in the past. With the significant increase in the number of training data sets and the dramatic increase in the processing power of chips, such as GPUs, deep learning has achieved outstanding results in the areas of target detection and computer vision, natural language processing, speech recognition and semantic analysis. Convolutional neural networks are one of the classic and widely used structures. CNN can automatically learn features from data through multi-layer nonlinear transformation, instead of manual design features. Since the milestone work of AlexNet [2], the novel CNN structure has significantly improved the accuracy of ImageNet classification, such as VGG [3], GoogLeNet [4], Inception-Resnet [5] ,ResNet [6,7], DenseNet [8], ResNeXt [9], SE-Net [10] and automatic neutral architecture search [11 -13].

It can be seen that the general trend of neural network design has been to find larger and deeper models to obtain better precision. The number of these parameters and the computational cost are very large, which is feasible only in a series of high-performance computing cards released by NVIDIA. However, because of the limited computing resources in the actual embedded platform, deep neural network model compression [14] including pruning, quantization, lightweight network model design, model tensor decomposition and approximate calculation are all indispensable steps. Among them, lightweight network model design is a very important study for deploying deep neural network on embedded systems.

To our best knowledge, in order to measure the computational complexity of a model, the widely used metrics are digital floating-point operations or FLOPs. However, FLOPs are indirect indicators. It is an approximation, but usually does not equal the direct metrics we really care about, such as speed or delay. Even the same FLOPs network runs differently in different platforms and code-optimized environments, so only deploying the network on the hardware platform to test the actual time is the real comparison standard. We leverage real hardware embedded platform to evaluate the lightweight model. In brief, we make the two main contributions in this paper: **(1)** Designed L-Mobilenet bottleneck in combination with Inception-ResnetV1 and MobilenetV2 [15], which significantly reduced parameters, memory access times and facilitated operation on embedded platforms. Experiments show that L-Mobilenet is faster on the GPU platform than the previous network, and the accuracy is comparable. (2) L-Mobilenet is deployed on the real ZYNQ platform and the network speed up is better than the baseline network.

## 2. RELATED WORK

The lightweight network model design is based on skill and experience to design a detailed, efficient and small-depth deep network model. This method mainly designs new network structures, such as the combination of network layers and changes in network connection relationships. Recently, including Squeezenet [16], Xception [17], MobileNet [18], ShuffleNet [19], MobileNet, DeepRebirth

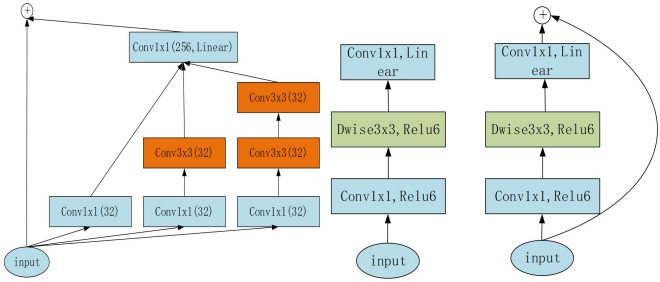

**Fig. 1**. Current lightweight neural network architecture, Inception-resnetv1 structure (left), Mobilenet_V2 (right)

[20], CondenseNet [21], ShufflenetV2[22], Squeezenext [23] led to these Better speed - accuracy trade-offs.

Among them, Mobilenet is a small convolutional neural network model designed by Google for embedded devices. The main design idea is to decompose the ordinary convolution kernel. In other words, the process can be interpreted as decompose the process of convolution kernel whose size is into two convolutions of and constitutes a deep separable convolution operation, which greatly reduces the amount of computation and only 1% loss on the ImageNet dataset compared to VGG-16.

ShufflenetV1 model take advantage of the channel information. The linear combination of the convolution process has a large amount of computation, and it is intended to reduce the amount of computation by group convolution [24] operations, but each group of the group convolution can only accept part of the information of the previous layer, and cannot integrate the previous network layer. All channels, therefore, the channel shuffle operation method, each convolution group can accept the data of different convolution groups of the previous network layer.

MobilenetV2 is an improved version of Mobilenet, the main idea is to use the reverse residual and the idea of channel expansion solves the problem that Mobilenet deep separable convolution has serious information loss after Relu activation in the case of less channels and the test time is optimal on the CPU.

DeepRebirth divided into weight layer (such as convolution layer and fully connected layer) and non-weight layer (such as Pooling layer, ReLU layer, etc.). The theoretical calculation of non-weight layer is small, but due to memory data access speed and other reasons, the computational cost is high. The method of combining the non-weighting layer and the weighting layer can remove the independent non-weighting layer, the running time is significantly reduced.

In the article ShufflenetV2, FLOPs are only theoretical approximations, not the measure of the network running speed. The real time is when the network runs on the actual embedded platform, and according to the experimental results of different networks on the ARM architecture side, it summarizes some basic operations that are not conducive to the operation of the embedded platform, such as Eltwise, ReLu and Add, etc. It is recommended, and based on this, to design an efficient network of speed and precision trade-off.

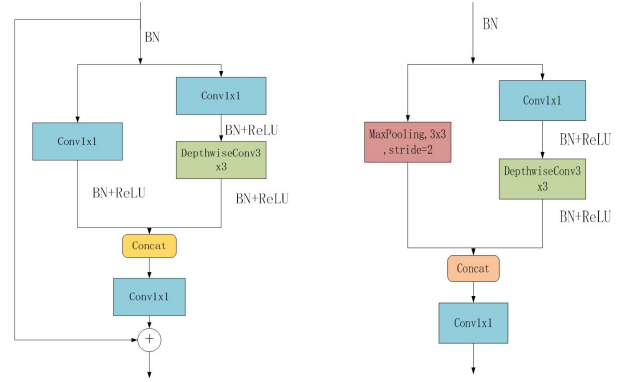

**Fig. 2.** L-Mobilenet Module, Stride=1 (left), Stride=2 (right)

Squeezenext is a model of software and hardware co-design, which believes that the depth separable convolution operation does not work well on some hardware platforms, so its block design level adopt low-rank convolution kernel and two-stage further compression channels, then dynamically adjust the number of layers of the network according to the time and power consumption results of hardware simulation results, etc. It can be seen that the continuous improvement of Block, in which the change of convolution mode and the new operation combination are the key to its design, both of which are small in calculation amount and the number of parameters is small for the purpose of hardware operation.

More and more attention to the efficiency of the network in the actual platform rather than just staying in the theoretical value to measure and compare. Software and hardware collaborative design and optimization is a more novel and practical approach, which is more conducive to making more realistic System. Therefore, Block and network architecture design, software and hardware cooperative tuning is the key and most practical path for lightweight deep neural network structure design.

## 3. BOTTLENECK DESIGN

### 3.1. Motivation: Inception-resnet_V1 and Mobilenet_V2

Figure 1 shows the Inception-resnetV1 and MobilenetV2 modules. The Inception-resnetV1 module uses different sizes of convolution kernels to enhance the width and adaptability of the network. The convolution kernels of 1*1 and 3*3 sizes are used to further reduce the number of parameters. The module of MobilenetV2 first expands the channel with a convolution kernel of size 1*1, and then mitigates feature degradation by increasing the InputChannels of ReLU. Inspired by the Inception-resnet-V1 and Mobilenet-V2, we find that the middle 3*3 depthwise convolution of MobilenetV2 can be combined with the Inception-ResnetV1 module, and the single module 3*3depthwise can be changed to 1*1 and 3*3 depthwise alternatively, which can achieve the effect of multi-scale extraction features while reducing the number of parameters.

### 3.2. L-Mobilenet Module and Design Consideration

Based on the hardware characteristic, the design principle of Bottleneck in L-Mobilenet can be summarized as five aspects:

**1)**. Taking advantage of the 1*1 and 3*3 convolution kernel parallel structure of Inception-ResnetV1. When the step size is one, the 1*1 convolution kernel of the left branch of the parallel structure in the left graph of Figure2 can be interpreted as two functions: (1) Increase the channels. (2) The 1*1 convolution kernel is followed by the Relu function to enhance the nonlinear characteristics of the network. (3) The 1*1 convolution kernel does not function as a feature extraction, so it can be seen as a feature reuse in a Bottleneck. 1*1 is mainly to expand the input channels number of 3*3 Depthwise; when the step size is 2, in order to reduce the size of the parameters and featuremap, we try to reduce the size of the featuremap by using AveragePooling and MaxPooling. Finally, we find that the left parallel structure uses 3*3 Maxpooling to reduce the featuremap size and maintain the translation invariance effect is the best. The size and the invariance of the featuremap are optimal and there are no extra parameters.

**2)**. According to the idea of ShufflenetV2: the memory access cost $MAC=h*w*(c1+c2)+c1*c2$; $B=h*w*c1*c2$ (the $h$ and $w$ are the length and width of the featuremap, respectively, $c1$ and $c2$ are the input channels and output channels of the 1*1 convolution kernel). The closer the $c1$ and $c2$ are, the better (the minimum value of $MAC$ is obtained when $c1 = c2$), and the expansion and compression of the channels are realized at the same time. The number of expansions of the two 1x1 convolution kernels was chosen to be doubled, which was also confirmed in different proportions of hardware experiments.

**3)**. L-Mobilenet module does not use group convolution which is first used in Alexnet. Although group convolution is beneficial to further reduce parameters and calculations, enhance network sparsity. On the contrary, according to ShufflenetV2, the group convolution operations increase the number of MAC.

**4)**. There is no more structure like the four branches of Googlenet's Inception or even other Inception series. We only use two parallel structures here, which have two advantages: (1) we can reduce the Cache miss rate as much as possible. The tensor in the layer Bottleneck can be decomposed into 6-way branch tensor calculations and then combined, so that although the Cache miss is increased due to the storage, the 2-5 will gradually decrease. (2) the shufflenetV2 indicates the network fragmentation ( For example, GoogLeNet's multipath structure) will reduce the degree of parallelism, so choose the least two parallel structures here.

**5)**. We have abandoned the practice of BN and ReLU in the previous Bottleneck structural design after the convolution layer, and borrowed the same variant residual unit proposed in Identity Mappings in Deep Residual Networks[7] to improve the accuracy.

### 3.3. Network Structure and Detailed Analysis

Fig. 3 shows the overall architecture of L-Mobilenet, which has 38 layers. The *Input* represents the size of the featuremap and the number of output channels for each layer. *Operator* represents the operation of the layer. *Expand* represents the channels relative to the input channels after Concat operation. The *C* represents the number of channels in each layer, and the *N* represents the operation repeated several times, *S* represents the step size. Here the expansion factor is selected 4 because the structure of 2. ResNet in Section 3 is actually not friendly to bandwidth. The calculation of the bypass is very small. The characteristics of the Eltwise operation are very large, so the bandwidth is relatively tight. Here, the number of channels is set less than the previous network. On the one hand, it is beneficial to hardware storage; on the other hand, it reduces memory. Increase the speed with the number of times/frequency of Cache access.

In addition to the convolution layer of feature extraction, there are other non-convolution layers that cannot be ignored in the convolutional neural network (Table1). The operation of these non-convolution layers also accounts for a considerable proportion of the running time of the convolutional neural network. DeepRebirth's non-tensor layer occupancy time histogram and ShufflenetV2's guidance suggestion (4) are mentioned here. Here we count the number of each operation of our network and MobilenetV2, ShufflenetV2 non-convolution layer, compared to MobilenetV2, BN ratio and ReLU, the number of Eltwise is reduced, the element operation is greatly reduced, although the 11 concat operations is more, concat is a non-element operation, which takes almost no time; compared to ShufflenetV2, the BN and ReLU functions are greatly reduced. Small, although the number of Eltwise is more than ShufflenetV2, we can see that the BatchNormalization layer takes the most time due to the time-consuming histogram in Deepbirth (this is also confirmed in later experiments), so our entire network running time is less than ShufflenetV2.

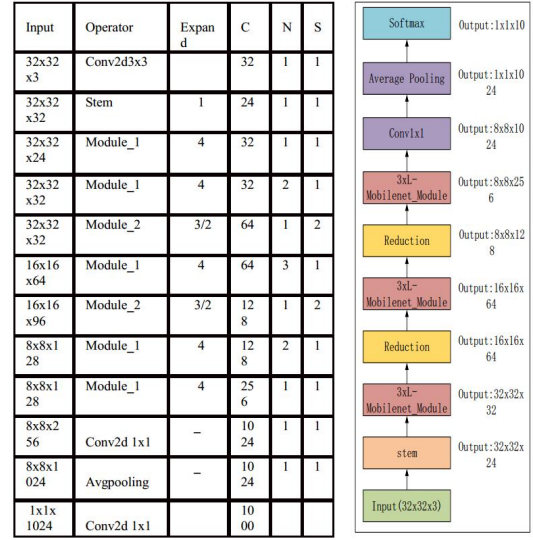

| Input | Operator | Expand | C | N | S |
|---|---|---|---|---|---|
| 32x32x3 | Conv2d3x3 | | 32 | 1 | 1 |
| 32x32x32 | Stem | 1 | 24 | 1 | 1 |
| 32x32x24 | Module_1 | 4 | 32 | 1 | 1 |
| 32x32x32 | Module_1 | 4 | 32 | 2 | 1 |
| 32x32x32 | Module_2 | 3/2 | 64 | 1 | 2 |
| 16x16x64 | Module_1 | 4 | 64 | 3 | 1 |
| 16x16x96 | Module_2 | 3/2 | 128 | 1 | 2 |
| 8x8x128 | Module_1 | 4 | 128 | 2 | 1 |
| 8x8x128 | Module_1 | 4 | 256 | 1 | 1 |
| 8x8x256 | Conv2d 1x1 | – | 1024 | 1 | 1 |
| 8x8x1024 | Avgpooling | – | 1024 | 1 | 1 |
| 1x1x1024 | Conv2d 1x1 | | 1000 | | |

**Fig 3**. L-Mobilenet architecture and architecture diagram

**Table 1.  Statistics of Non-convolution layer operations**

| Network | BatchNormalization | ReLu | Eltwise | Concat |
|---|---|---|---|---|
| L-Mobilenet | 46 | 35 | 7 | 11 |
| MobilenetV2 | 54 | 36 | 10 | 0 |
| ShufflenetV2 | 56 | 37 | 0 | 16 |

### 3. EXPERIMENTS

To fully evaluate our design, we run our CNN model on different hardware platforms. The processor used includes GPU, ARM and ZYNQ:

-**GPU:** A single NVIDIA GeForce GTX TITAN Xp with is used.The driver version is 384.130, the environment is cuda8.0, and the platform is Tensorflow1.6 [26]. The Cudann acceleration is not used.

-**ARM:** ZC702 and ZC706: Double Cortex-A9,which is based on ARMv7-A architecture;ZCU102:Quad-core ARM® Cortex-A53,which is based on ARMv8-A architecture.

**GPU Results: (Shown in Table 2)**

All models are trained using back-propagation [27] by Stochastic Gradient Descent [28] with Nesterov momentum [29] (NAG) optimizer implemented by MXNet for 320 epochs.The initial learning rate is set to 0.1 and is reduced 10 times at 150 and 225 epochs The parameters are initialized by Xaviers initializer [30]. The other settings are:weight decay of 0.0001, momentum of 0.9, and batch size of 64. (one TiTanxp GPU)), where the network of 1.0.

MobilenetV2 is in the output channel The 64 and 160 settings have a step size of 2, the remaining steps are all one, the ShufflenetV2 1× Stage2 has a step size of 1, and the Stage3 and Stage4 steps are set to 2. The rest are one. The actual parameters in Table 2 represent the size of all parameters including the BN layer.

| Table 2. Experiment results on GPU | | | | | |
|---|---|---|---|---|---|
| Neural network | Images/s | Params (Theoretical value) | Params(Actual value) | Cifar10 | Cifar100 |
| L-Mobilenetv2 | 1250 | 0.9M | 19.7M | 0.9344 | 0.7500 |
| Mobilenetv2 | 650 | 3.4M | 48M | 0.94 | 0.7633 |
| Shufflenetv2 | 960 | 2.3M | 27M | 0.9223 | 0.7209 |
| PydMobilenet-56-0.75 | 1180 | 0.489M | 5.7M | 0.9295 | 0.6895 |

**Embedded system results: (Shown in Table 3, 4)**

1)All networks are implemented in embedded C language and run on pure ARM side in bare metal case. The results are as shown in Table 3.

| Table 3. Experiment results on ZYNQ with programmable syetem | | | |
|---|---|---|---|
| | ZC702 | ZC706 | ZCU102 |
| L-Mobilenet | 15.2s | 15.2s | 1.11s |
| MobilenetV2 | 41.8s | 41.6s | 3.34s |
| ShufflenetV2 | 26.9s | 26.9s | 2.36s |

2) After Accelerating with ZYNQ programmable logic

We leverage the Xilinx SDSoC tools to accelerate the C code in ARM processor. The TCF Profiler in SDSoc can track threads and count the proportion of each function occupying ARM resources. When the network runs on the ZYNQ platform, by analyzing the resource occupancy ratio of each function on the ARM side, we use the acceleration code of SDSoC to accelerate the highest resource occupation. The result is shown in Table 4.

| Table 4. Experiment results on ZYNQ with programmable logic (PL) | | | |
|---|---|---|---|
| | ZC702 | ZC706 | ZCU102 |
| L-Mobilenet | 15.1s | - | - |
| MobilenetV2 | 41.7s | - | - |
| ShufflenetV2 | 24.5s | - | 2.15s |

When leveraging the SDSoC to accelerate the ARM side code, the upper layers of the network of L-Mobilenet and Mobilenetv2 occupy the highest level of ARM resources. We tried the acceleration of the featuremap layer of 8*8, 16*16, 32*32 size, we found that the 16*16 size are moved to the PL side and run longer than the only ARM side, because the number of convolutional channels with output sizes of 8*8 and 16*16 are increased. More total cycle times lead to an increase in ARM-side scheduling time, even though the PL-side convolution operation takes less time, but it also inspires a network that is designed to facilitate the acceleration operations.

### 4. CONCLUSION

We propose a lightweight neural network for embedded and mobile terminals called L-MobileNet, a new model architecture. It significantly reduces parameters and computational delays, compared to the current popular network. L-Mobilenet is deployed in ZYNQ embedded system with programmable logic (PL). The platform can run our network and accelerate the code in ARM side. Experimental results show that our model can achieve better performance as the model has more hardware consideration.